\pdfoutput=1

\documentclass[11pt]{article}

\usepackage[]{latex/acl}

\usepackage{times}
\usepackage{latexsym}

\usepackage[T1]{fontenc}

\usepackage[utf8]{inputenc}

\usepackage{microtype}

\usepackage{inconsolata}

\usepackage{listings}

\usepackage{amsmath}
\usepackage{amsfonts}
\usepackage{amssymb}
\usepackage{graphicx}
\usepackage{color}
\usepackage{url}
\usepackage[boxed,vlined]{algorithm2e}
\usepackage{multirow}
\usepackage{multicol}

 \newcommand{\bit}{\begin{itemize}}
 \newcommand{\eit}{\end{itemize}}
 \newcommand{\ben}{\begin{enumerate}}
 \newcommand{\een}{\end{enumerate}}

\newcommand{\method}{\textsc{LIM-A}\xspace}
\newcommand{\methodr}{\textsc{LIM-RA}\xspace}

\newcommand{\alignscore}[0]{AlignScore\xspace}
\newcommand{\roberta}[0]{RoBERTa\xspace}
\newcommand{\deberta}[0]{DeBERTa\xspace}
\newcommand{\distillbert}[0]{DistilBERT\xspace}

\newcommand{\robustname}[0]{Robust-Name\xspace}
\newcommand{\robustnum}[0]{Robust-Number\xspace}

\newcommand{\llmr}[0]{LLMR\xspace}

\usepackage{titlesec}

\usepackage{caption}

\newcommand{\sref}[1]{Section~\ref{#1}} 
\newcommand{\fref}[1]{Figure~\ref{#1}} 
 
\newcommand{\tref}[1]{Table~\ref{#1}}

\usepackage{boldline}
\usepackage{pifont}%
\usepackage{amsmath,amstext}
\usepackage{tablefootnote}

\usepackage{colortbl}

    \usepackage{flushend}

\title{Less is More for Improving Automatic Evaluation of Factual Consistency}

\author{ Tong Wang, Ninad Kulkarni, Yanjun Qi \\
   AWS Bedrock Science \\
   \texttt{\{tonwng, ninadkul, yanjunqi\}@amazon.com}}

\begin{document}
\maketitle

\begin{abstract}

Assessing the factual consistency of automatically generated texts in relation to source context is crucial for developing reliable natural language generation applications. Recent literature proposes AlignScore which uses a unified alignment model to evaluate factual consistency and  substantially outperforms previous methods across many benchmark tasks. In this paper, we take a closer look of datasets used in AlignScore and uncover an unexpected finding: utilizing a smaller number of data points can actually improve performance. We process the original AlignScore training dataset to remove noise, augment with robustness-enhanced  samples, and utilize a subset comprising 10\% of the data to train an improved factual consistency evaluation model, we call LIM-RA (\texttt{Less Is More for Robust AlignScore}). LIM-RA demonstrates superior performance, consistently outperforming AlignScore and other strong baselines like ChatGPT across four benchmarks (two utilizing traditional natural language generation datasets and two focused on large language model outputs).  Our experiments show that LIM-RA achieves the highest score on 24 of the 33 test datasets, while staying competitive on the rest, establishing the new state-of-the-art benchmarks.

\end{abstract}

\begin{figure}[t]
    \centering
    \includegraphics[width=0.5\textwidth]{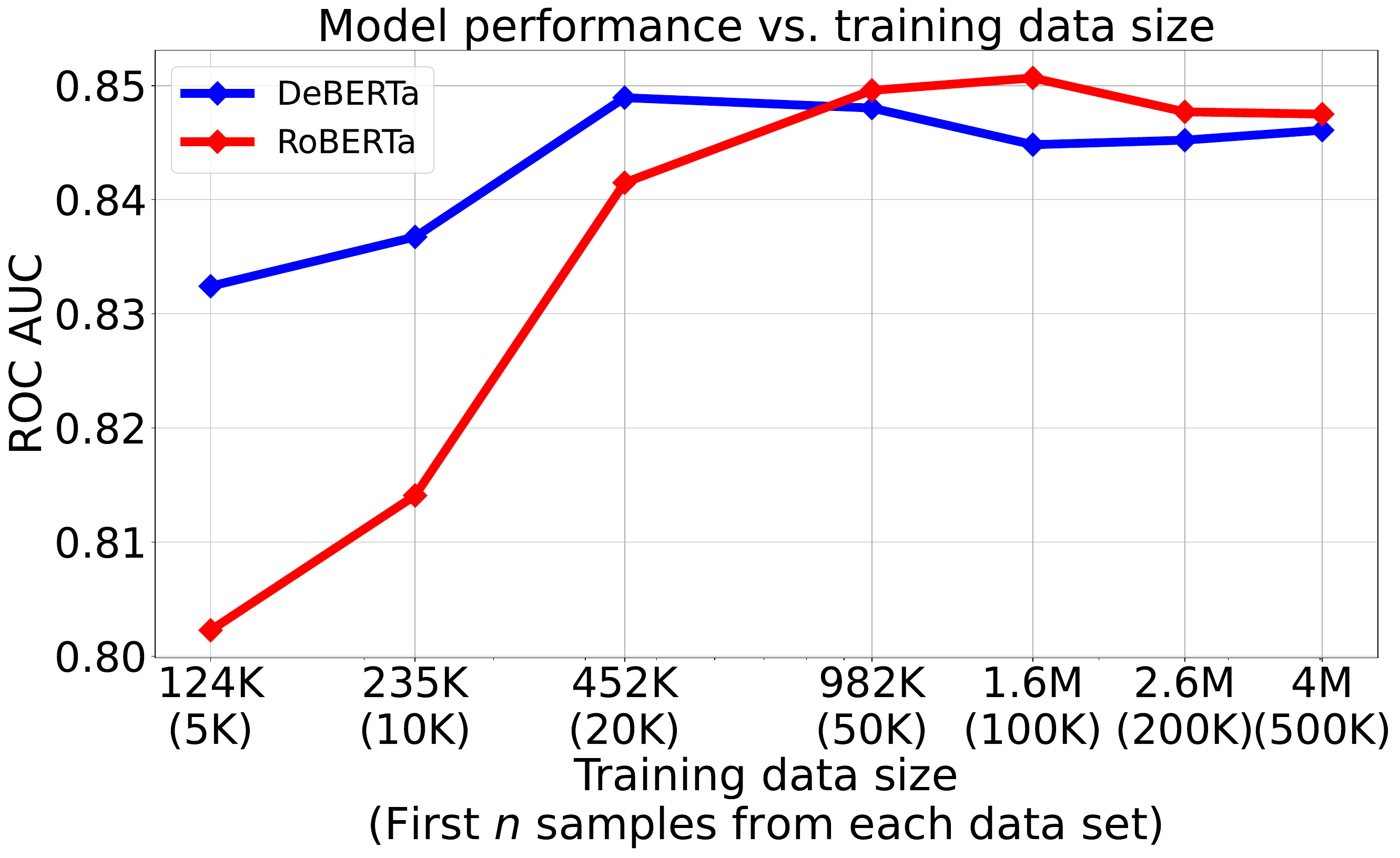}
	\caption{Ablation study on using the first $n$ samples from each sub-train dataset for training and overall model performance. We see that the optimum benchmark performance is 452K and 1.6M samples for DeBERTa and RoBERTa respectively. For comparison \alignscore uses 4.7M or the first 500K. Performance broken down by benchmark can be found in \ref{sec:appendix_ablation}}
	\label{fig:training_data}
\end{figure}

\section{Introduction}

The emergence of large language models (LLMs) and an increasing interest in utilizing machine-generated texts from like summarization,  paraphrasing, and question-answering (QA) has created a need to automatically evaluate  the degree to which generated natural language texts accurately reflect the factual information contained in source context. Early work used Natural Language Inference (NLI) \cite{laban2022summac} and QA \cite{fabbri2021qafacteval} to handle automatic factual consistency evaluation. However, these methods exhibit limited generalizability and struggle with handling long contexts. Recently, \citet{zha2023alignscore} propose \alignscore, a unified model based on RoBERTa and is trained on a wide range of datasets to calculate the alignment between context and generated text. \alignscore achieves state-of-the-art results across several factual consistency benchmarks.

Despite its strengths, the \alignscore study has several limitations.  First, the training data leveraged for developing \alignscore were derived in a heuristic manner from many existing NLP tasks and datasets, adding noise and poor quality in some samples. We, therefore, ask the question: Are all data points from \alignscore training needed? Our ablation studies shown in Figure~\ref{fig:training_data} indicate that the answer is "No". Additionally, \alignscore displays fragility regarding robustness, as it fails to identify some clear perturbations involving entities like names, numbers, etc. As Table~\ref{tab:alignscore_errors} illustrates, even simple modifications can produce false positives and false negatives  when using \alignscore.

\begin{figure*}
    \small
    \centering
	\includegraphics[width=15cm]{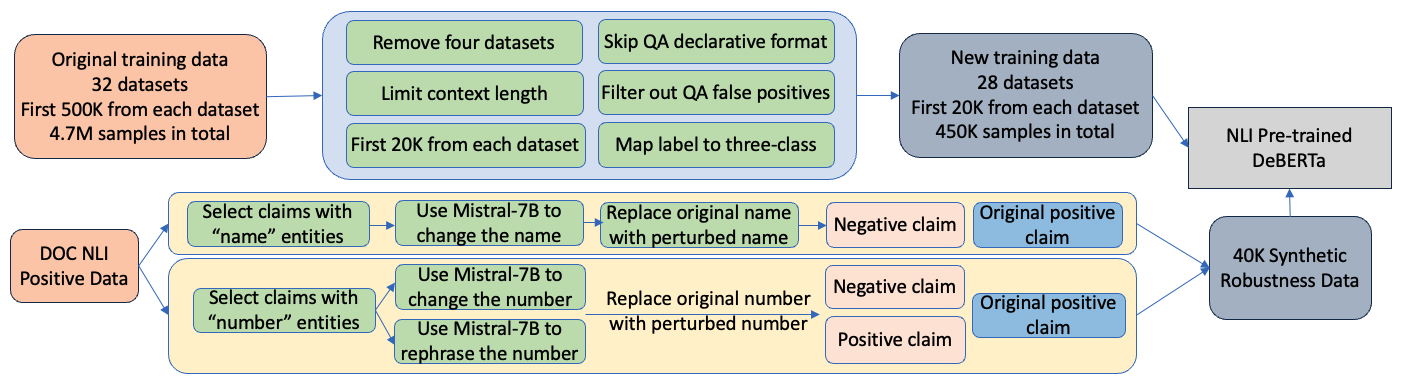}
	\caption{Overall workflow of our method is depicted in the diagram. The top workflow describes how we clean the training data, the bottom workflow illustrates the process of creating synthetic robustness data. Then we train a pre-trained \deberta model on those data to obtain \methodr.}
	\label{fig:overall_workflow}
\end{figure*}

 In this paper, we propose \textbf{\methodr} (\textbf{L}ess \textbf{I}s \textbf{M}ore - \textbf{R}obust \textbf{A}lignScore), an improved version of AlignScore trained on \deberta ~\cite{he2021debertav3}. Our model is the result of multiple ablation steps on improving training data quality, analyzing training size as well as constructing synthetic data to improve robustness (\fref{fig:overall_workflow} shows overall workflow). We demonstrate that with about 10\% of the cleaned training data, we are able to obtain a better model than \alignscore.  Our experiments show that \methodr consistently outperforms strong baselines including \alignscore and GPT-3.5-Turbo, achieving the new state-of-the-art  on four factual consistency benchmarks covering a wide range of 33 datasets. It is worth noting that our experiments include a newly defined benchmark, Large Language Model Response (LLMR), designed for evaluating LLM outputs' factual consistency. \methodr performs the best on LLMR.

\section{Method}

\begin{table}[!tp]
  \small
  \centering
  \scalebox{0.9}{
  \begin{tabular}{p{0.23\linewidth} | p{0.39\linewidth} | p{0.17\linewidth} | p{0.1\linewidth}}
    \hline
    Context & Claim & \alignscore & \methodr \\
    \hline
    \multirow{2}{=}{[...] Napoleon married the Archduchess \textcolor{blue}{Marie Louise}, who was 18 years old [...]} & Archduchess \textcolor{blue}{Marie Louise} was 18 years old when she married Napoleon . &  0.9907 & 0.9542 \\ \cline{2-4}
     & Archduchess \textcolor{red}{Mari Louze} was 18 years old when she married Napoleon . & 0.9650  (\textcolor{purple}{false positive}) & 0.4381 \\
    \hline
    \multirow{2}{=}{The Blue Ridge Mountains [...] attain elevations of about \textcolor{blue}{2,000} ft} & The typical elevations of the Blue Ridge Mountains are \textcolor{blue}{2,000} ft. &  0.9812 & 0.9434\\ \cline{2-4}
     & The typical elevations of the Blue Ridge Mountains are \textcolor{blue}{2000} ft. & 0.0214 (\textcolor{purple}{false negative}) & 0.8621 \\
    \hline
  \end{tabular}
  }
  \caption{Examples of robustness issues in \alignscore predictions. In the first example we perturb the correct name "Marie Louise" to the incorrect name "Mari Louze"; however, the factual consistency score is still high, resulting in a false positive. Similarly in the second example we perturb  "2,000" to "2000", resulting in a false negative.}
  \label{tab:alignscore_errors}
 \end{table}

\subsection{\alignscore Model and Training Data}

Automatic evaluation of factual consistency  is challenging. Recently proposed \alignscore measures the alignment of information between machine-generated natural language texts and given source material to evaluate the factual accuracy \cite{zha2023alignscore}.  
\alignscore is built on top of a unified alignment function via \roberta ~\cite{liu2019roberta} and trained on datasets derived from 7 NLP tasks: NLI, QA, Fact Verification, Paraphrase, Semantic Textuality Similarity, Information Retrieval, and Summarization. Each sample in a task is converted into a text pair (context, claim) and a label. The label has 3 options based on the task and dataset: binary (aligned, not-aligned), 3-way (aligned, contradict, neutral), regression (score between 0 to 1). For example in SNLI dataset, the context is the premise, the claim is the hypothesis, label is the 3-way label. 
Certain prepossessing steps are required to unify the format in multiple datasets. 

To calculate the factual consistency score of long text, \alignscore first splits the context into roughly 350-token chunks and the claim into sentences. Then the trained alignment function (\roberta based) evaluates each sentence in the claim against each context chunk. For example, in the 3-way classification head, the probability of the "aligned" class is used as the alignment score. The highest alignment score for each claim sentence is selected and then averaged to obtain the overall factual consistency score. By using the chunking strategy, \alignscore can be applied to text of any length, as shown by \fref{fig:alignscore_diagram}.

\subsection{Training Data Cleaning}
For training, \alignscore uses more than 30 datasets and selects 500K samples from each dataset to build its trainng data, including a total of 4.7M training samples. Training the \alignscore alignment model requires 5 days on 8 V100 GPUs.

However, we find that not all the training datasets have good quality. The upper half of \fref{fig:overall_workflow} shows a cohort of data cleaning steps we use to improve the training data quality. First, based on our ablation studies, we remove four datasets that do not result in performance gains, such as ms\_marco and wikihow. Additionally to prevent the model from truncating sentences that support the claim, we only keep samples in which the context has fewer than 512 tokens.

When using QA datasets to create alignment training samples, since the QA passage is the context, a preprocessing step is needed. \alignscore uses a pre-trained sequence-to-sequence model to convert question-answer into a declarative sentence as the input claim. We, however, observed a performance decrease in our experiments when using this preprocessing. We find the decrease was because the generated declarative sentence has poor data quality. Thus, we concatenate question and answer as the claim text. \footnote{We also tried to use Mistral-7B \cite{jiang2023mistral} few-shot to generate better-quality declarative sentences but still did not produce performance gains.} 

Additionally, many QA datasets only have ground truth answers (positive samples) but no wrong answers (negative samples). To address this, \alignscore generates fake wrong answers using the T5 model, and answers the question based on the original passage with the ground truth answer tokens masked. However, this leads to false negatives because many generated fake answers are similar to or exactly match their corresponding ground truth answers. To mitigate the issue,  we use Sentence-BERT \cite{reimers2019sentence} to encode both the fake and ground truth answers, and then filter out the fake answers that are similar to the true answers by using rules and a threshold of 0.85. This data cleaning procedure is illustrated in the top half of figure~\ref{fig:overall_workflow}.

After cleaning the data, we use 20K samples from each dataset for a total of 452K training samples (about 10\% of  training data used for \alignscore) which results in a better model (results in \sref{exp:result}).

\subsection{Synthetic Robustness Data}
\label{sec:synthetic_data}

We also notice \alignscore fails on name or number perturbations as illustrated in Table \ref{tab:alignscore_errors}. To mitigate the issue, we augment the training dataset by creating a synthetic dataset designed to enhance the model's robustness, with emphasis on name and number variation based text generation as illustrated in the bottom half of figure~\ref{fig:overall_workflow}.

We create two synthetic datasets: \robustname and \robustnum datasets using DocNLI ~\cite{yin2021docnli}. DocNLI includes multiple-sentence contexts and single-sentence claims discussing facts in the context. To create the \robustname data, we use spaCy NER ~\cite{spacy2} to identify the "PERSON" and "ORG" entities in samples labeled as "entailment" and use Mistral-7B to perturb the entities (prompt details in Appendix~\ref{sec:appendix_synthetic_few_shot}). The original entity is replaced with the perturbed entity to construct the synthetic negative samples. Using Mistral instead of randomly perturbing a character in the entity ensures the new name is similar to a real person or org name. The two-step generation generates a better rewritten claim than directly instructing the LLM to rewrite the claim.

Similarly, we construct the \robustnum data by perturbing claims with number-related labels such as "TIME", "QUANTITY", "DATE". We use Mistral to rephrase ("100" to "one hundred") and change numbers ("100" to "101"). The perturbed entities replace the original to create positive and negative data.

\subsection{\methodr Model}
We experiment with different pretrained models as base including \roberta (large), \deberta (large), \distillbert (base). \deberta achieves the best overall performance while \distillbert has poor performance due to its small model capacity. Also, %
we unify all data labels to the three class setup (details later in this section %
), and use the 3-way classification head to predict $aligned$ (factual consistent), $neutral$ (no-evidence), and $contradiction$. At inference time, we follow \alignscore to split context into chunks and claim into sentences, and average the sentence alignment scores to compute the overall factual consistency score. We denote \methodr and \method as the \deberta model trained with cleaned data and with and without synthetic robustness in training, respectively.

\textbf{Under the Hood:} We train a pre-trained NLI \deberta model\footnote{\textit{https://huggingface.co/MoritzLaurer/DeBERTa-v3-large-mnli-fever-anli-ling-wanli}} ~\cite{laurer2024less} for 3 epochs using AdamW optimizer with learning rate as 1e-5. We use the first 20k samples from each of the 28 train datasets described in \alignscore, plus the 2 new synthetic robustness datasets, resulting in a total of 490k samples in our final training. Hyperparameter details can be found in Table~\ref{sec:appendix_hyperparameters}. We follow \alignscore and use the factual consistency class probability as the alignment score.

\paragraph{Unifying Labels:}
\label{sec:appendix_unify_label}
We convert binary and regression labels to 3-class labels. For datasets with binary labels, we map the negative label ``not-aligned" to either ``contradiction" or ``no-evidence" depend on the dataset. In most of the cases, we map the negative label to ``contradiction", such as in \textit{doc\_nli} and \textit{paws}. But in \textit{qqp}, we map the negative label to ``no-evidence". For regression labels in \textit{stsb} dataset, we bin the score as three classes: faithful ($>=0.45$), no-evidence ($>=0.3, <0.45$), contradiction ($<0.3$).

\subsection{Connecting to Related Works}

Previous studies include multiple other methods for assessing factual consistency. 
(1) QA-based factual consistency, including QuestEval \cite{scialom2021questeval} and QAFactEval \cite{fabbri2021qafacteval}, checks if the source answer is different from the target answer given a question. (2) With the recent advances in LLMs, a new line of research is to evaluate factual consistency directly with an LLM \cite{liu2303g, fu2023gptscore,jia2023zero}. \cite{chen2023evaluating} investigate a variety of prompting methods including vanilla prompting, chain-of-thought prompting, and a sentence-by-sentence prompting and \cite{luo2023chatgpt} explore ChatGPT’s ability to evaluate factual inconsistency under a zero-shot setting while \cite{fu2023large} uses LLMs in a QA setting for direct factual consistency scoring. (3) A third related line of methods uses the Natural Language Inference (NLI) based formulation. For instance \cite{laban2022summac} proposed SummaCConv, that segments documents into sentences and aggregates NLI scores between pairs of sentences.

Factual consistency benchmark datasets typically contain (context, claim, label) triplets where the label indicates if the claim is consistent with the context and is difficult to obtain as high-quality annotation is challenging due to low inter-annotator agreement \cite{falke2019ranking, laban2022summac}. \cite{laban2022summac} introduce the SummaC (Summary Consistency) benchmark which consists of 6 large inconsistency detection datasets standardized as a binary classification task given document and summary. %
\cite{laban2023llms} introduce SummEdits, a summarization consistency dataset where an LLM introduces inconsistencies in an otherwise consistent summary and show that the benchmark is challenging for most current LLMs. \cite{honovich2022true} present TRUE, which consolidates 11 existing datasets covering summarization, knowledge-grounded dialogue, paraphrasing and fact verification annotated for consistency.

\begin{table}[t]
  \centering
  \small
  \begin{tabular}{ccccccc}
    \hline
    Model & CG & XF & FC & SE & FRK & AVG \\
    \hline
    NER         & 54.4          & 69.0       & 50.8      & 59.3     & 68.4     & 60.4 \\
    Questeval   & 59.7          & 65.6       & 73.3      & 76.9     & 86.3     & 72.4 \\
    QAFactEval  & 82.5         & 65.1       & 89.2      & 88.5     & 89.6     & 82.9 \\
    SummaC      & 65.6          & 70.3      & 92.2   & 86.0     & 88.4     & 80.5 \\
    \alignscore & 76.9          & \textbf{78.1} & 89.1      & 82.3     & 88.0     & 82.9 \\
    \method     &84.2	        & 73.9	 & \textbf{93.7}	& \textbf{92.4}	& 90.3    & 86.0 \\
    \methodr     & \textbf{84.9}          & 75.7       & 93.2      & 92.2     & \textbf{92.0}     & \textbf{87.6} \\
    \hline
  \end{tabular}
  \caption{SummaC benchmark AUC-ROC results. \methodr and \method outperform all current baselines in 4 of the 5 datasets with \methodr performing the best overall.}
  \label{tab:summac_results}
 \end{table}

 \begin{table*}[!ht]
  \centering
  \small
  \begin{tabular}{cccccccccccc}
    \hline
    Model & ECT & QM & SCall & SS & SCI  & SEmail & NEWS & BILL & PD & SP & AVG \\
    \hline
    NER         & 59.7 & 55.2 & 56.3 & 57.6 & 53.3 & 66.8 & 60.9 & 49.1 & 57.8 & 51.5 & 56.8 \\
    Questeval   & 64.8 & 54.0 & 63.1 & 54.4 & 51.9 & 55.9 & 64.9 & 59.8 & 54.1 & 53.4 & 57.6 \\
    QAFactEval  & 75.8 & 65.5 & 74.6 & 71.3 & 69.7 & 69.8 & 81.4 & 56.9 & 64.0 & 65.5 & 69.4 \\
    SummaC      & 66.7 & 55.8 & 61.1 & 54.3 & 61.0 & 58.9 & 61.1 & 54.5 & 61.0 & 61.1 & 59.6 \\
    \alignscore & 91.5 & 83.8 & 89.1 & \textbf{85.5} & 82.1 & 81.6 & 80.6 & 61.6 & 78.0 & 72.3 & 80.6 \\
    \method   & \textbf{93.6} & 86.9 & 90.7 & 83.1 & \textbf{87.7} & \textbf{82.5} & \textbf{82.1} & 69.3 & 81.0 & 86.9 & 84.4 \\
 
    \methodr     & 92.8 & \textbf{88.2} &  \textbf{91.2}	& 84.2 & 86.0 & 81.1 &	81.3 &	\textbf{72.1}	& \textbf{82.5}	& \textbf{86.9} & \textbf{84.6} \\
    
    \hline
  \end{tabular}
  \caption{SummEdits benchmark AUC-ROC results. \methodr and \method are the best performing models in 9 of the 10 datasets and \methodr is the best performing model overall.}
  \label{tab:summedits_results}
 \end{table*}

 \begin{table*}[htb]
  \centering
  \small
  \begin{tabular}{ccccccccc}
    \hline
    Model & ATS & BBA-4 & BBA-16 &BBS-4 &BBS-16 &PHD	&HE & AVG \\
    \hline
    \alignscore  &62.7 & 62.4 & 59.4 & 71.8 & 75.2 & 74.6 & 73.1 & 68.5\\
    \method &65.9  & 69.2 & 60.4& 79.5& 78.4 & \textbf{78.0} & 72.2 & 71.9 \\
    \methodr  & \textbf{66.3} & \textbf{71.4} & \textbf{60.7} & \textbf{82.5} & \textbf{79.6} & 77.0 & \textbf{74.9} & \textbf{73.2} \\
    \hline
  \end{tabular}
  \caption{\llmr benchmark AUC-ROC results. We compare against only \alignscore as it is the best performing baseline as seen in tables~\ref{tab:summac_results} and \ref{tab:summedits_results}. \methodr and \method are the best performing models in all 7 datasets. \methodr is the best performing model overall.}
  \label{tab:llm_response_results}
 \end{table*}

 \begin{table*}[!htb]
  \centering
  \small
  \begin{tabular}{cccccccccccccc}
    \hline
    Model & BEGIN & DF & FVR &FRK &MNBM &PAWS	&Q$^2$	&QC	&QX	&SE	& VITC & AVG$^*$ & AVG\\
    \hline
    NER        & 50.6 & 62.7 & 62.4 & 65.5 & 68.4 & 51.7 & 59.1 & 48.4 & 63.6 & 56.6 & 57.8 & 59.3 & 58.8\\
    Questeval & \textbf{83.9} & 77.2 & 72.5 & 84.0 & 64.8 & 69.0 & 72.2 & 64.5 & 55.2 & 69.7 & 66.6 & 71.4 & 70.9\\
    QAFactEval   & 81.0 & 81.8 & 86.0 & 88.5 & 67.3 & 86.1 & 75.8 & 83.9 & 76.1 & 80.9 & 73.6 & 79.4 & 80.1\\
    SummaC   & 81.6 & 81.2 & 92.0 & 89.0 & 67.2 & 88.2 & 77.5 & 77.7 & 76.0 & 79.1 & 97.5 & 78.7 & 82.5\\
    \alignscore & 81.4 & 85.0 & 94.9 & 88.7 & \textbf{78.2} & 98.3 & 79.1 & \textbf{89.6} & \textbf{83.1} & 71.4 & \textbf{98.4} & 82.0 & 86.2 \\
    \method & 79.0 & \textbf{85.2} & \textbf{95.5} & 90.0 & 74.7 & 98.4 & \textbf{83.5} & 85.0 & 82.4 & 83.5 & 97.1 & 83.0 & 86.8\\

    \methodr & 80.8 & 83.8 & 95.2 & \textbf{91.3} & 75.8 & \textbf{98.4} & 82.7 & 84.5 & 82.7 & \textbf{84.8} & 96.8 & \textbf{83.3} & \textbf{87.0} \\
    
    \hline
  \end{tabular}
  \caption{TRUE benchmark AUC-ROC results. \methodr and \method are the best performing models in 6 of the 11 datasets. \methodr is the best performing model overall. We
    report AVG$^*$ in the second last column by excluding PAWS, FVR, and VITC to show out-of-domain performance.}
  \label{tab:true_results}
 \end{table*}

 \begin{table*}[htb]
  \centering
  \small
  \begin{tabular}{llllll}
    \hline
    Model & SummaC & TRUE & SummEdits & LLMR & AVG \\
    \hline
    \alignscore & 82.9 & 86.2 & 80.6 & 68.5 & 79.6 \\
    \method & 86.0 (+3.7\%) & 86.8 (+0.7\%) & 84.4 (+4.7\%) & 71.9 (+5.0\%) & 82.3 (+3.4\%)\\
    \methodr & \textbf{87.6 (+5.7\%)}  & \textbf{87.0 (+0.9\%)} & \textbf{84.6 (+5.0\%)} & \textbf{73.2 (+6.9\%)} & \textbf{83.1 (+4.4\%)} \\    
    \hline
  \end{tabular}
  \caption{Average AUC results and relative improvements over \alignscore on four benchmarks. The last column is the overall average of SummaC, TRUE, SummEdits, and LLMR scores.}
  \label{tab:all_results}
 \end{table*}

\begin{table}[t]
  \centering
  \small
  \begin{tabular}{cccc}
    \hline
     Model & Setting & Overall \\
    \hline
    \alignscore & 4.7M & 83.2 \\
    \hline
    \roberta & pre-train & 71.6 \\
    \deberta & pre-train & 82.1 \\
    \hline
    \roberta & 10\% + cleaning & 84.1 \\
    \deberta & 10\% + cleaning & 83.6 \\
    \hline
    \roberta & 10\% + cleaning + pre-train & 83.8  \\    
    \method & 10\% + cleaning + pre-train & 86.0 \\
    \hline
    \methodr & +syn robust data & 86.4\\
    \hline
  \end{tabular}
  \caption{Ablation Study}
  \label{tab:ablation_study}
  \vspace{-5mm}
 \end{table}

\section{Experiments}

We conduct a comprehensive experimental study to evaluate \methodr on multiple factual consistency benchmarks and demonstrate \methodr consistently outperforms strong baselines and establishes new state-of-the-art results. Our experiments also include ablation studies (\tref{tab:ablation_study}) and robustness analysis (\tref{tab:synthetic_results}) of \methodr. We list the hyperparameters we used for \methodr in Table~\ref{tab:hyperparameters}. Each of our experiments covers 20 different random seeds.

\subsection{Four Benchmarks: 33 Datasets}
We evaluate the factual consistency performance using AUC-ROC on 33 datasets from 4 benchmarks: SummaC, SummEdits, TRUE, and \llmr. Each data sample in the benchmarks is a pair of target text (claim) and a grounding source text (context), with a binary annotation of whether the target text is factually consistent w.r.t its source. The benchmark dataset details can be found in Appendix~\ref{sec:appendix_benchmark}.

\textbf{SummaC} 5 summary consistency datasets: GoGenSumm (CG), XsumFaith (XF), FactCC (FC), SummEval (SE), Frank (FRK). We remove Polytope dataset since it contains negative samples that do not imply factual consistency errors.

\textbf{TRUE} 11 datasets covering summarization, knowledge-grounded dialogue, paraphrasing and fact verification annotated for factual consistency: Frank (FRK), SummEval (SE), MNBM, QAGS-CNNDM (QC), QAGS-Xsum (QX), BEGIN, Q$^2$, DialFact (DF), Fever (FVR), VitaminC (VITC), PAWS.

\textbf{SummEdits} 10 datasets evaluating factual consistency in summarization covering multiple domains. Inconsistent summaries are generated by GPT-3.5-Turbo: News, Podcast (PD), Billsum (BILL), Samsum (SS), Shakespeare (SP), SciTLDR (SCI), QMSum (QM), ECTSum (ECT), Sales Email (SEmail), Sales Call (SCall).

\textbf{\llmr} (large language model response) is a new benchmark consisting of 7 datasets we introduce in this paper. Similar to SummEdits, the datasets are designed to evaluate the factual consistency of LLM output and inconsistencies are generated in an automated fashion with human verification: HaluEval (HE) \cite{HaluEval} consists of CNN/DailyMail articles with correct and hallucinated summaries generated by ChatGPT in a zero-shot manner. BAMBOO abs-hallu (BBA) and sen-hallu (BBS) subsets \cite{dong2023bamboo} consist of NLP academic papers (max 4K and 16K token variants for a total of 4 datasets) with supported and hallucinated hypotheses generated by ChatGPT similar to HE. Passage-level Hallucination Detection (PHD) \cite{yang-etal-2023-new-benchmark} consists of Wikipedia articles of an entity with correct and hallucinated biographies of that entity generated by ChatGPT. AttrScore (ATS) \cite{yue2023automatic} consists of QA datasets and New Bing search queries in the format $(question, answer, context, label)$ where $label$ indicates if the $answer$ is supported by $context$. Hallucinations are generated by both swapping the answer with an incorrect answer and by swapping the the context with another article. For our experiments we consider context as $document$ and answer as $claim$.

\subsection{Baselines Methods}

\textbf{NER} ~\cite{laban2022summac}, uses spaCy NER to match entities between claim and context.

\noindent\textbf{Questeval}, QA-based model, evaluates both factual consistency and relevance of the generated text by checking if the answer from source is different from the answer from target given a question. 

\noindent\textbf{QAFactEval}, QA-based model, evaluates factual consistency by performing answer selection, question generation, question answering, and answer overlap evaluation.

\noindent\textbf{SummaC}, NLI-based model (SummaCConv), segments documents into sentence units and aggregates scores between pairs of sentences.

\noindent\textbf{\alignscore}, current state-of-the-art, an alignment function trained on a wide range of datasets.

\noindent\textbf{0-shot/10-shot GPT-3.5-Turbo}, instruct the LLM to evaluate whether the claim is consistent, lacks evidence, or contains contradictions.

\noindent\textbf{10-shot Mistral-7B}, one of the best performing open-source LLMs. We use the same prompts as 10-shot GPT-3.5-Turbo.

\subsection{Experimental Results}
\label{exp:result}

\subsubsection{Results on Traditional Benchmarks: SummaC and TRUE}

We evaluate factual consistency models on the SummaC benchmark in Table~\ref{tab:summac_results}. \methodr achieves the best overall score and has a 5.7\% relative improvement over \alignscore and QAFactEval. Our model has the top result in 4 of the 5 datasets. Our results for \alignscore are lower than the results reported in the original work~\cite{zha2023alignscore} because we did not include the rule-based inference-time processing (such as removing special tokens or capitalizing the first letter) for a fair comparison between all models. 

From the results on the TRUE benchmark in Table~\ref{tab:true_results}, we see that \methodr has the best overall AUC-ROC score with a 0.9\% improvement over \alignscore and has the best score in 5 of 11 datasets. As suggested in ~\cite{zha2023alignscore}, we report AVG$^*$ by removing PAWS, FVR, and VITC to show out-of-domain performance; \methodr remains the best performing model.

\subsubsection{Results on LLM output: SummEdits and \llmr}
\label{sec:llm_response_result}

We evaluate factual consistency on LLM responses using the SummEdits and \llmr benchmarks in Table~\ref{tab:summedits_results} and Table~\ref{tab:llm_response_results} respectively. On the SummEdits benchmark, both \method and \methodr consistently outperform other baselines. \methodr has the best overall performance and has a 5.0\% relative improvement over the best baseline \alignscore. Our model achieves the best score in 8 of the 10 datasets and performs significantly better on OOD domain datasets such as Shakespeare (SP), BillSum (BILL), SciTLDR (SCI) compared to the baseline. On the \llmr benchmark, we only report \alignscore as Tables~\ref{tab:summac_results},~\ref{tab:summedits_results},~\ref{tab:true_results} show that \alignscore is the strongest baseline. \methodr achieves the best overall result and obtains a relative improvement of 6.9\% over \alignscore, and has the best score on 6 of the 7 datasets.

We report the overall average score on the four benchmarks in Table~\ref{tab:all_results}. In summary, \methodr exhibits a 4.4\% relative improvement over the baseline model \alignscore.

\subsubsection{Comparing with LLM Baselines}
\label{sec:llm_baseline}

We compare the trained metric models with two LLMs: Mistral-7B and GPT-3.5-Turbo (ChatGPT) using the same 0-shot and 10-shot prompt (described in Appendix~\ref{sec:appendix_evaluator_few_shot}). Since LLMs do not provide factual consistency scores, we report balanced accuracy in Table~\ref{tab:llm_results} and only report SummaC and SummEdits due to time constraints. \methodr continues to perform the best on the two benchmarks while GPT-3.5-Turbo outperforms Mistral by a large margin on SummaC. Additionally, 0-shot ChatGPT outperforms 10-shot ChatGPT on SummEdits possibly because the 10-shot demonstrations are out-of-domain. We compare average inference time of each model on a sample of data from SummaC and find \alignscore demonstrates fast inference speed of 0.18s on a single NVDIA-A10G GPU followed by \methodr with 0.29s. The slower speed is because \deberta is slower than \roberta even though they have a similar number of parameters. 0-shot ChatGPT and Mistral-7B on 4 GPUs using vLLM~\cite{kwon2023efficient} achieves comparable speed of 0.52s and 0.51s respectively while OpenAI GPT-3.5 10-shot is the slowest, primarily due the to the rate limit of a Tier-1 account\footnote{The Tier-1 rate-limit for GPT-3.5-Turbo is 60K tokens per minute, 3.5K requests per minute, and 10K requests per day. https://platform.openai.com/docs/guides/rate-limits/usage-tiers?context=tier-one}.

\begin{table}[t]
  \centering
  \small
  \begin{tabular}{ccccc}
    \hline
    Model & SC & SE & Time & GPUs \\
    \hline
    Mistral-7B 10-shot  &62.0 &64.0 & 0.51s & 4 \\
    GPT-3.5 0-shot & 73.5 & 71.6 & 0.52s & API \\
    GPT-3.5 10-shot &76.7 & 69.8 & 8.4s & API \\
    \alignscore  &74.0 &71.9 & 0.18s & 1  \\
    \method  & 77.8  & 76.5 & 0.29s & 1 \\
    \methodr  & \textbf{78.5}  & \textbf{76.7} & 0.29s  & 1 \\
    \hline
  \end{tabular}
  \caption{Evaluation using LLMs Balanced Accuracy results and Average Inference Time on SummaC (SC) and SummEdits (SE).}
  \label{tab:llm_results}
 \end{table}

\subsection{Results on Synthetic Robustness Data}

In Table~\ref{tab:synthetic_results} we evaluate the models on the synthetic robustness test dataset created in section~\ref{sec:synthetic_data}. We see \method without synthetic data augmentation performs on par with \alignscore while \methodr performs the best and is more robust to name and number perturbations.

 \begin{table}[t]
  \centering
  \small
  \begin{tabular}{ccc}
    \hline
     & \robustname & \robustnum  \\
    \hline
    Train (Test) & 19,508 (3,492) & 20,628 (5,076) \\
    \hline
    \alignscore  & 64.3 &  86.4 \\
    \method  & 64.0  &  88.0 \\
    \methodr  & \textbf{84.8}  & \textbf{91.8}  \\
    \hline
  \end{tabular}
  \caption{Synthetic robustness data size and AUC-ROC performance across models when facing perturbed data.}
  \label{tab:synthetic_results}
 \end{table}

\subsection{Ablation Analysis}
\label{sec:abl}

We perform ablation studies to answer the following questions: (1) What is the impact of different training data sizes? (2) What is the performance of using a pre-trained model as the alignment? (3) What is the impact of the cleaned data? and (4) What is the impact of fine-tuning \roberta or \deberta as the alignment function? 

To answer (1) we sweep the size from 123K (5K per dataset) to 4M (500K per dataset). From Figure~\ref{fig:training_data}, we see that the benchmark performance peaks at 452K and 1.6M samples for DeBERTa and RoBERTa respectively and reduces if we include more data. For (2)-(4), we report the average AUC-ROC score of SummaC, SummEdits, TRUE in Table \ref{tab:ablation_study}. To answer (2), we experiment with different off-the-shelf pre-trained NLI models. The best pre-trained \deberta model (82.1\%) outperforms the best pre-trained \roberta \footnote{\textit{https://huggingface.co/ynie/roberta-large-snli\_mnli\_fever\_anli\_R1\_R2\_R3-nli}} (71.6\%)~\cite{nie2019adversarial}. To answer (3), we perform data cleaning and use 10\% (452K samples) of the training data and find both \roberta (84.1\%) and \deberta (83.6\%) outperform \alignscore (83.2\%). To answer (4), we fine-tune the pre-trained models using the cleaned data. \deberta (\method) performance improves with fine-tuning while \roberta performance decreases, possibly because the pre-trained \deberta outperforms the pre-trained \roberta model. Finally, adding the synthetic robustness data can further boost the performance.

\section{Conclusions}
\label{sec:concl}

We propose \methodr, a \deberta based model to automatically evaluate factual consistency trained from a cleaner and smaller training set than used for \alignscore. Experimental results show \methodr consistently outperforms the current state-of-the-art \alignscore and other strong baselines on 4 benchmarks. In addition, the model is robust to name and number variations and is better suited for LLM outputs' factual consistency evaluation.

\bibliography{refNLP23,reffaithfulness}

\clearpage
\newpage
\appendix
\begin{figure}
    \small
    \centering
	\includegraphics[width=6cm]{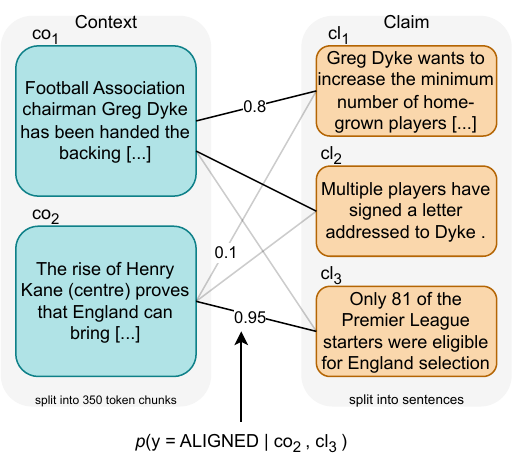}
	\caption{Visual description of \alignscore. The context and claim are split into 350 token and sentence chunks respectively. Then an alignment function evaluates each \textit{(context chunk, claim sentence)}. The factual consistency score is calculated by first selecting the highest alignment score for each \textit{claim} and then averaging these scores across all \textit{claims}.}
	\label{fig:alignscore_diagram}
\end{figure}

\section{Appendix}
\label{sec:appendix}

\subsection{Training data size ablation for each benchmark dataset}
\label{sec:appendix_ablation}
\begin{figure}[!h]
    \centering
    \includegraphics[width=0.5\textwidth]{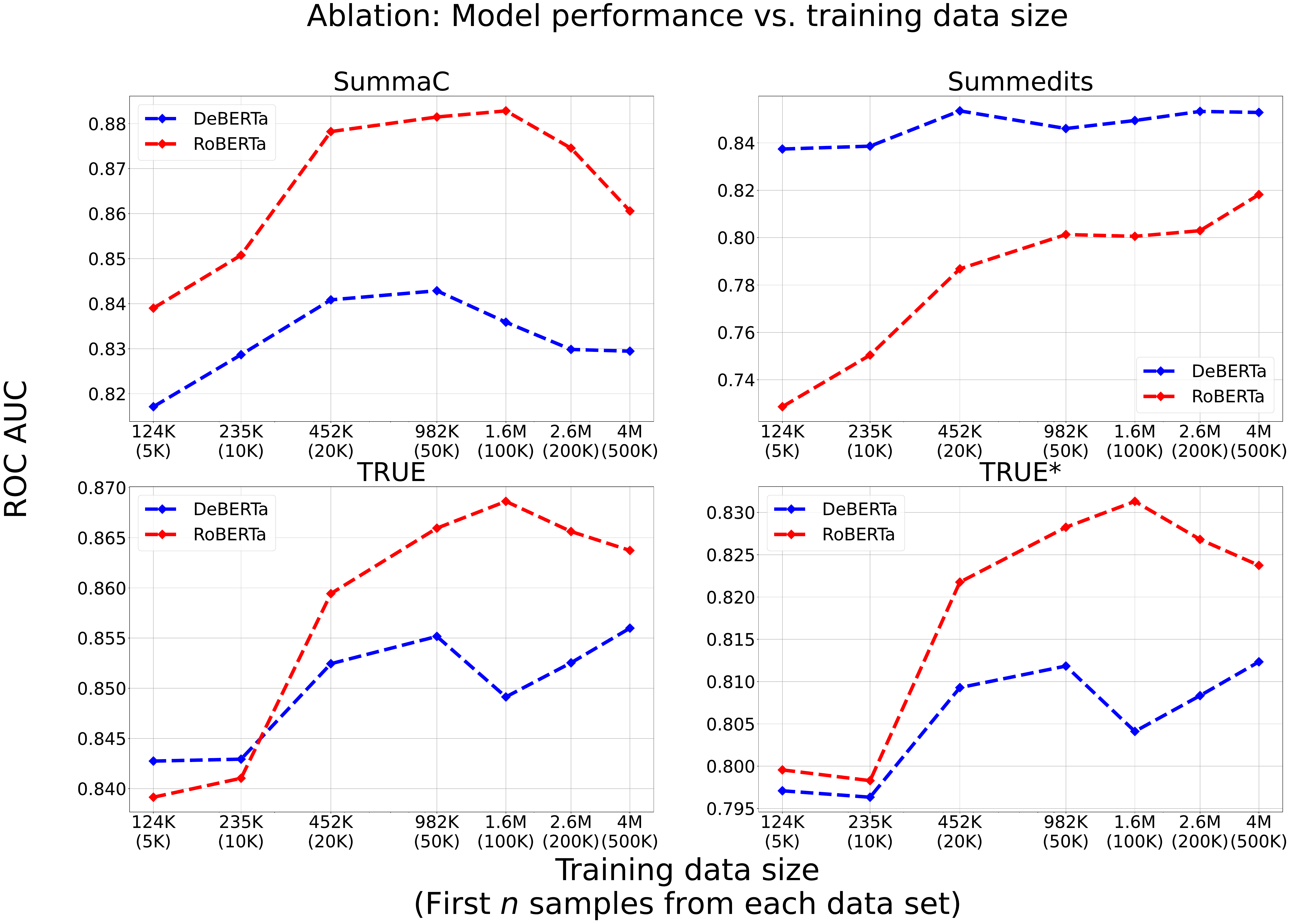}
	\caption{Ablation study on using the first $n$ samples for training and model performance on each benchmark data set. }
	\label{fig:training_data_broken_down}
\end{figure}

\begin{table}[!h]
  \centering
  \small
  \begin{tabular}{l|l}
    \hline
    Parameter & Value  \\
    \hline
    samples\_per\_dataset & 20000 \\
    max\_context\_length & 512 \\
    lr & 1e-5 \\
    seed & 2027 \\
    train\_batch & 8 \\
    accumulate\_grad\_batch & 1 \\
    epoch & 3 \\
    warmup\_ratio & 0.06 \\
    weight\_decay & 0.01 \\
    adam\_epsilon & 1e-6 \\
    \hline
  \end{tabular}
  \caption{\methodr Hyperparameters}
  \label{tab:hyperparameters}
  \label{sec:appendix_hyperparameters}
 \end{table}

\subsection{Benchmark Details}
\label{sec:appendix_benchmark}
Tables~\ref{tab:summac_benchmark}, \ref{tab:summedits_benchmark}, \ref{tab:true_benchmark}, \ref{tab:llmr_benchmark} describe the total number and number of factually consistent samples in each benchmark and dataset.

\begin{table}[!h]
  \centering
  \small
  \begin{tabular}{l|ll}
    \hline
    Dataset & \# Samples & \# Factually Consistent  \\
    \hline
    CG & 400 & 312 \\
    XF & 1250 & 130 \\
    FC & 503 & 441 \\
    SE & 850 & 770 \\
    FRK & 1575 & 529 \\
    \hline
    Total & 4578 & 2182 \\
    \hline
  \end{tabular}
  \caption{SummaC Benchmark}
  \label{tab:summac_benchmark}
 \end{table}

 \begin{table}[!h]
  \centering
  \small
  \begin{tabular}{l|ll}
    \hline
    Dataset & \# Samples & \# Factually Consistent  \\
    \hline
    ECT & 668 & 242 \\
    QM & 431 & 183 \\
    SCall & 520 & 173 \\
    SS & 664 & 242 \\
    SCI & 466 & 145 \\
    SEmail & 613 & 179 \\
    NEWS & 819 & 321 \\
    BILL & 853 & 361 \\
    PD & 500 & 163 \\
    SP & 814 & 378 \\
    \hline
    Total & 6348 & 2387 \\
    \hline
  \end{tabular}
  \caption{Summedits Benchmark}
  \label{tab:summedits_benchmark}
 \end{table}

 \begin{table}[!h]
  \centering
  \small
  \begin{tabular}{l|ll}
    \hline
    Dataset & \# Samples & \# Factually Consistent  \\
    \hline
    BEGIN & 836 & 282 \\
    DF & 8689 & 3341 \\
    FVR & 18209 & 6393 \\
    FRK & 671 & 223 \\
    MNBM & 2500 & 255 \\
    PAWS & 8000 & 3539 \\
    Q$^2$ & 1088 & 628 \\
    QC & 235 & 113 \\
    QX & 239 & 116 \\
    SE & 1600 & 1306 \\
    VITC & 63054 & 31484 \\
    \hline
    Total & 105121 & 47680 \\
    \hline
  \end{tabular}
  \caption{TRUE Benchmark}
  \label{tab:true_benchmark}
 \end{table}

 \begin{table}[!h]
  \centering
  \small
  \begin{tabular}{l|ll}
    \hline
    Dataset & \# Samples & \# Factually Consistent  \\
    \hline
    ATS & 4241 & 1414 \\
    BBA-4 & 200 & 100 \\
    BBA-16 & 200 & 100 \\
    BBS-4 & 200 & 100 \\
    BBS-16 & 200 & 100 \\
    PHD & 299 & 222 \\
    HE & 20000 & 10000 \\
    \hline
    Total & 25340 & 12036 \\
    \hline
  \end{tabular}
  \caption{LLMR Benchmark}
  \label{tab:llmr_benchmark}
 \end{table}

\subsection{Few-shot Prompt to Generate Synthetic Robustness Data}
\label{sec:appendix_synthetic_few_shot}

\subsubsection{Prompt to perturb names}
\begin{lstlisting}
Given a name, modify one or two 
letters to change it to a different
name.

Original Text: Abraham Lincoln
Changed Text: Abrahem Lincoln

Original Text: cricket
Changed Text: cracket

Original Text: Wireshark
Changed Text: Wileshark

Original Text: Robert Urquhart.
Changed Text: Robert Uruhart.

Original Text: Dee Smith
Changed Text: Dee Smyth

Original Text: Emma Wastson
Changed Text:
\end{lstlisting}

\subsubsection{Prompt to perturb numbers}
\begin{lstlisting}
Change the meaning of the text.

Original Text: 37
Changed Text: 27

Original Text: more than 10 
years ago
Changed Text: more than 11 
years ago

Original Text: more than 10 
years ago
Changed Text: within 10 years

Original Text: second
Changed Text: third

Original Text: 22 June 1990
Changed Text: 22 July 1990

Original Text: at least one
Changed Text: at most one

Original Text: 2 years
Changed Text:
\end{lstlisting}

\subsubsection{Prompt to rephrase numbers}
\begin{lstlisting}
Rephrase the numbers in the text.

Original Text: 154
Rephrase Text: one hundred 
fifty-four

Original Text: more than 10 
years ago
Rephrase Text: more than ten 
years ago

Original Text: second
Rephrase Text: 2nd

Original Text: 22 June 1990
Rephrase Text: June twenty-two 
nineteen ninety

Original Text: at least one
Rephrase Text: at lest 1

Original Text: twenty-five
Rephrase Text: 25

Original Text: 2001
Rephrase Text: two thousand and 1

Original Text: 2 years
Changed Text:
\end{lstlisting}

\subsection{Few-shot Prompt for Evaluating Factual consistency}
\label{sec:appendix_evaluator_few_shot}
\begin{lstlisting}
Decide if the claim is faithful 
with the corresponding context. 
Note that Factual consistency 
means all information 
in the claim is supported 
by the context. 
Answer with 0 (consistent), 
1 (no evidence), or 
2 (contradiction).

Context: I burst through a set 
of cabin doors, and fell to 
the ground-
Claim: I burst through the 
doors and fell down.
Answer: 0

Context: Fun for adults and 
children.
Claim: Fun for only children.
Answer: 2

Context: Thebes held onto power 
until the 12th Dynasty, when 
its first king, Amenemhet I who 
reigned between 1980 1951 b.c. 
established a capital 
near Memphis.
Claim: The capital near Memphis 
lasted only half a century 
before its inhabitants 
abandoned it for the 
next capital.
Answer: 1

[...]
\end{lstlisting}

\end{document}